\documentclass[lettersize,journal]{IEEEtran}
\usepackage{amsmath,amsfonts}
\usepackage{algorithmic}
\usepackage{algorithm}
\usepackage{array}
\usepackage[caption=false,font=normalsize,labelfont=sf,textfont=sf]{subfig}
\usepackage{textcomp}
\usepackage{stfloats}
\usepackage{url}
\usepackage{verbatim}
\usepackage{graphicx}
\usepackage{cite}
\usepackage{tikz}
\usetikzlibrary{calc}
\usepackage[colorlinks=true,
            linkcolor=blue,
            citecolor=blue,
            urlcolor=blue,
            bookmarks=true,
            bookmarksnumbered=true,
            pdftitle={Safety-Critical Camera Reliability Monitoring for ADAS},
            pdfauthor={Shiva Aher}]{hyperref}
\begin{document}
\pagestyle{empty}

\title{Safety-Critical Camera Reliability Monitoring for ADAS via Degradation-Aware Uncertainty Pattern Analysis}

\author{Shiva~Aher, Senior Member, IEEE%
\thanks{Shiva Aher is with the Georgia Institute of Technology, Atlanta, GA, USA. E-mail: saher8@gatech.edu.}%
\thanks{arXiv preprint, May 2026. Under review at IEEE Transactions on Intelligent Vehicles.}}

%\markboth{IEEE Transactions on Intelligent Vehicles}%
%{Aher: Safety-Critical Camera Reliability Monitoring for ADAS}

\maketitle
\thispagestyle{empty}

\begin{abstract}
Reliable camera input is essential for safety-critical ADAS perception, but most monitoring approaches detect sensor failures only after downstream performance has degraded. 
We propose a proactive camera reliability monitoring framework that estimates perception risk from degradation-induced uncertainty patterns before downstream failure becomes 
observable. The method introduces a Global Sensor Health Index (GSHI), a continuous reliability score that aggregates per-degradation severities using a risk-aware multiplicative 
formulation, allowing severe single-mode failures such as lens occlusion or motion blur to dominate the health estimate. A lightweight multi-task network predicts degradation type, 
severity, GSHI, and spatial uncertainty maps from a single RGB image without downstream task feedback. Training uses physics- and geometry-aware synthetic supervision over twelve 
camera degradation modes. Experiments on KITTI-derived degradations show that GSHI decreases monotonically with severity, achieves a health-estimation MAE of 0.064, and provides 
positive early-warning lead time of 0.47 ± 0.25 severity units before YOLOv8 detection failure. GSHI also outperforms IQA, detector-confidence, and clean-feature OOD baselines, 
and transfers zero-shot to real adverse-weather driving data. These results support degradation-aware uncertainty analysis as a practical direction for proactive camera reliability 
monitoring in intelligent vehicles.
\end{abstract}


\begin{figure*}[!t]
\centering

% consistent number formatting for labels
\newcommand{\GSHI}[1]{\scriptsize\textbf{GSHI: #1}}

\makebox[\textwidth][c]{%
\begin{tikzpicture}[x=1cm,y=1cm, font=\small]

% ---- Teaser layout auto-fit to \textwidth ----
\pgfmathsetmacro{\GapX}{0.10}
\pgfmathsetmacro{\GapY}{0.50}
\pgfmathsetmacro{\TileH}{1.55}
\pgfmathsetmacro{\LabelPad}{0.80}
\pgfmathsetmacro{\LeftPad}{0.00}
\pgfmathsetmacro{\TopPad}{0.00}
\pgfmathsetmacro{\NumRows}{3}

% Convert \textwidth from pt to cm
\pgfmathsetmacro{\TextWcm}{\textwidth/28.45274}

% Available width for 5 columns
\pgfmathsetmacro{\AvailW}{\TextWcm - \LabelPad - \LeftPad}
\pgfmathsetmacro{\TileW}{(\AvailW - 4*\GapX)/5.0}
\pgfmathsetmacro{\MapW}{\TileW}

% ---- X positions ----
\pgfmathsetmacro{\XStart}{\LeftPad + \LabelPad}
\pgfmathsetmacro{\XColClear}{\XStart}
\pgfmathsetmacro{\XColSFour}{\XStart + 1*(\TileW+\GapX)}
\pgfmathsetmacro{\XColSEight}{\XStart + 2*(\TileW+\GapX)}
\pgfmathsetmacro{\XColSOne}{\XStart + 3*(\TileW+\GapX)}
\pgfmathsetmacro{\XColMap}{\XStart + 4*(\TileW+\GapX)}

% ---- Column centers ----
\pgfmathsetmacro{\XMidClear}{\XColClear + 0.5*\TileW}
\pgfmathsetmacro{\XMidSFour}{\XColSFour + 0.5*\TileW}
\pgfmathsetmacro{\XMidSEight}{\XColSEight + 0.5*\TileW}
\pgfmathsetmacro{\XMidSOne}{\XColSOne + 0.5*\TileW}
\pgfmathsetmacro{\XMidMap}{\XColMap + 0.5*\MapW}

% ---- Y positions ----
\pgfmathsetmacro{\YHeader}{\NumRows*(\TileH+\GapY) + \TopPad - 0.22}
\pgfmathsetmacro{\YRowTop}{2*(\TileH+\GapY)}
\pgfmathsetmacro{\YRowMid}{1*(\TileH+\GapY)}
\pgfmathsetmacro{\YRowBot}{0*(\TileH+\GapY)}

% ---- Row labels ----
\pgfmathsetmacro{\XRowLabel}{\LeftPad + 0.18}
\pgfmathsetmacro{\YLabelTop}{\YRowTop + 0.5*\TileH}
\pgfmathsetmacro{\YLabelMid}{\YRowMid + 0.5*\TileH}
\pgfmathsetmacro{\YLabelBot}{\YRowBot + 0.5*\TileH}

% ---- Outer box ----
\pgfmathsetmacro{\XLeftBox}{\XStart - 0.05}
\pgfmathsetmacro{\YBotBox}{-0.05}
\pgfmathsetmacro{\XRightBox}{\XColMap + \MapW + 0.05}
\pgfmathsetmacro{\YTopBox}{\NumRows*(\TileH+\GapY) - \GapY + 0.05}

% ---- Styles ----
\tikzset{
  tile/.style={
    draw,
    rounded corners=1pt,
    minimum width=\TileW cm,
    minimum height=\TileH cm,
    inner sep=0pt
  },
  tilemap/.style={
    draw,
    rounded corners=1pt,
    minimum width=\MapW cm,
    minimum height=\TileH cm,
    inner sep=0pt
  }
}

% ---- Headers ----
\node[anchor=south] at (\XMidClear,\YHeader) {\textbf{Clear}};
\node[anchor=south] at (\XMidSFour,\YHeader) {\textbf{$s=0.4$}};
\node[anchor=south] at (\XMidSEight,\YHeader) {\textbf{$s=0.8$}};
\node[anchor=south] at (\XMidSOne,\YHeader) {\textbf{$s=1.0$}};
\node[anchor=south] at (\XMidMap,\YHeader) {\textbf{Uncertainty}};

% ---- Row labels ----
\node[anchor=center, rotate=90] at (\XRowLabel,\YLabelTop) {\textbf{Fog}};
\node[anchor=center, rotate=90] at (\XRowLabel,\YLabelMid) {\textbf{Snow}};
\node[anchor=center, rotate=90] at (\XRowLabel,\YLabelBot) {\textbf{Motion blur}};

% ---- Helpers: cover-crop without distortion ----
\newcommand{\ImTile}[5]{%
  \node[tile, anchor=south west] (#1) at (#2,#3) {};
  \begin{scope}
    \clip (#1.south west) rectangle (#1.north east);
    \node[anchor=center, inner sep=0pt] at (#1.center)
      {\includegraphics[height=\TileH cm]{#4}};
  \end{scope}
  \node[anchor=north] at ($(#1.south)+(0,-0.07)$) {#5};
}

\newcommand{\MapTile}[4]{%
  \node[tilemap, anchor=south west] (#1) at (#2,#3) {};
  \begin{scope}
    \clip (#1.south west) rectangle (#1.north east);
    \node[anchor=center, inner sep=0pt] at (#1.center)
      {\includegraphics[height=\TileH cm]{#4}};
  \end{scope}
}

% =========================
% Row 1: Fog / Haze
% =========================
\ImTile{FogClear}{\XColClear}{\YRowTop}{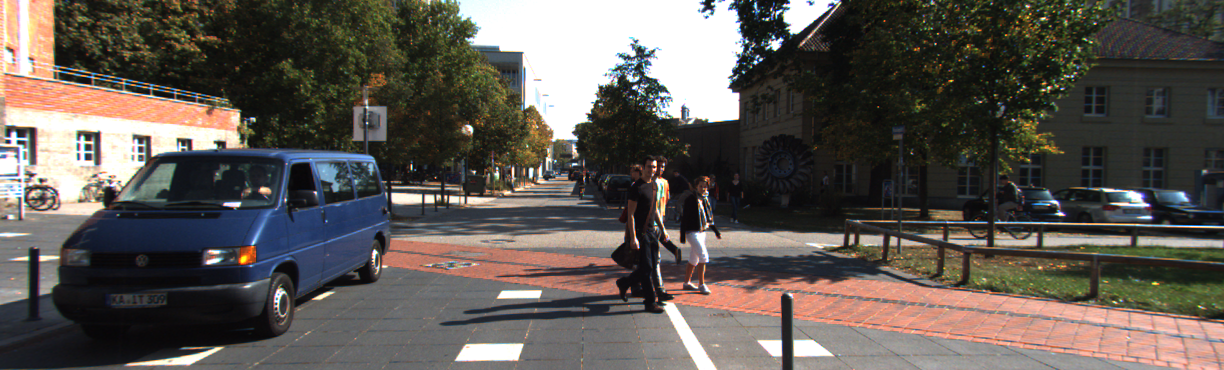}{\GSHI{0.819}}
\ImTile{FogFour}{\XColSFour}{\YRowTop}{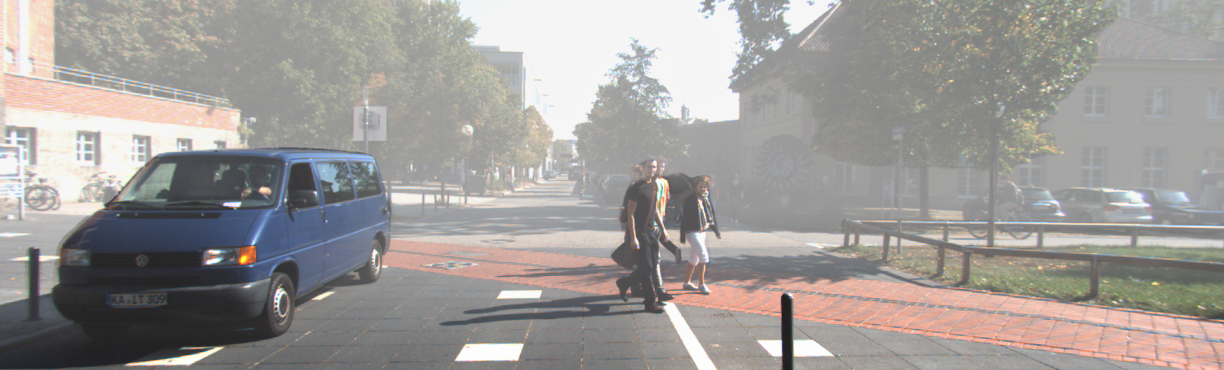}{\GSHI{0.543}}
\ImTile{FogEight}{\XColSEight}{\YRowTop}{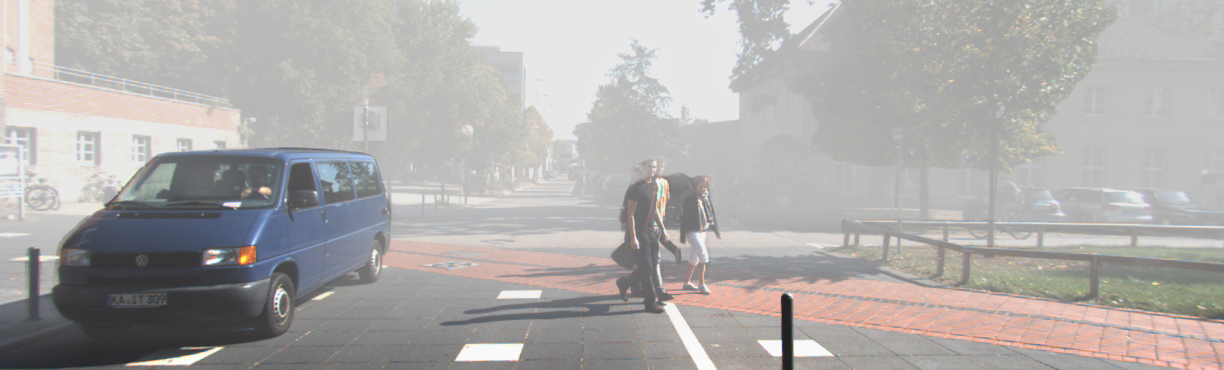}{\GSHI{0.218}}
\ImTile{FogOne}{\XColSOne}{\YRowTop}{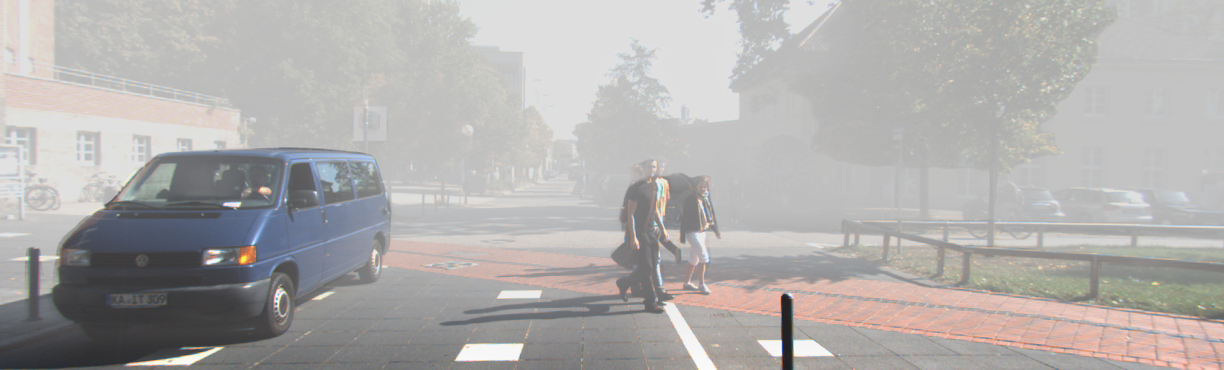}{\GSHI{0.171}}
\MapTile{FogMap}{\XColMap}{\YRowTop}{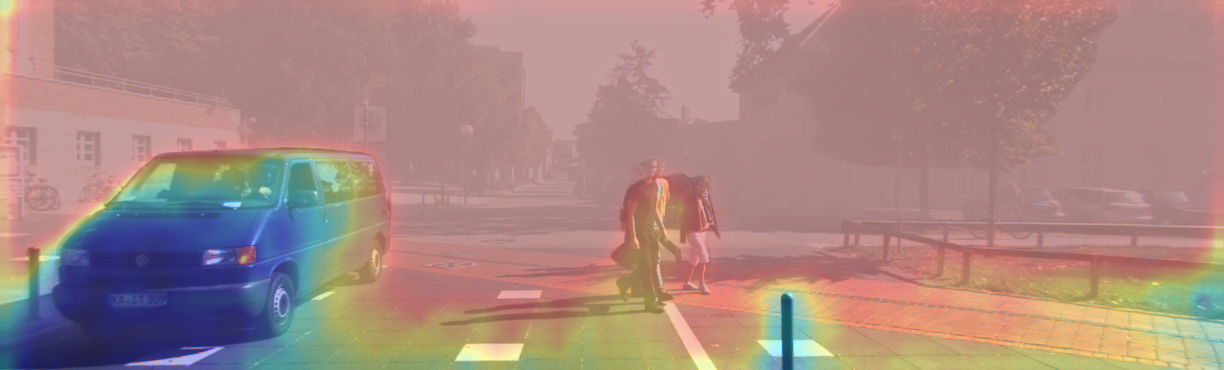}

% =========================
% Row 2: Snow
% =========================
\ImTile{SnowClear}{\XColClear}{\YRowMid}{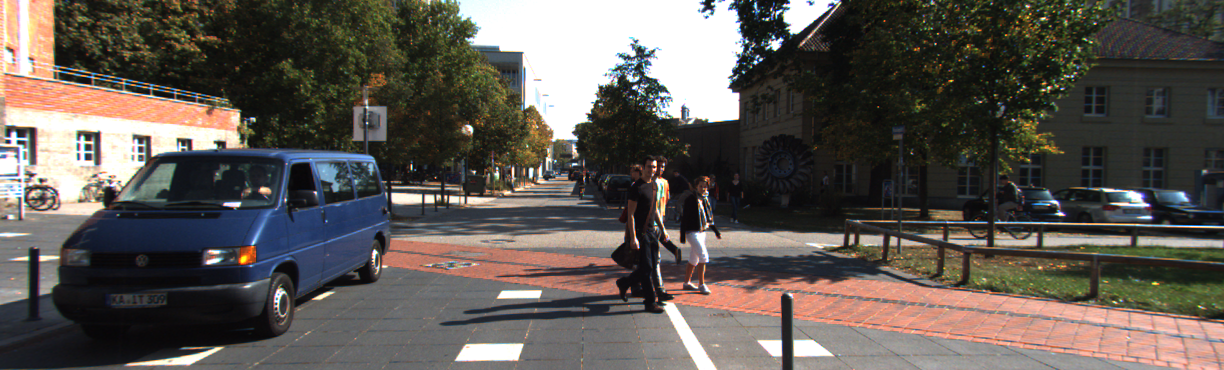}{\GSHI{0.819}}
\ImTile{SnowFour}{\XColSFour}{\YRowMid}{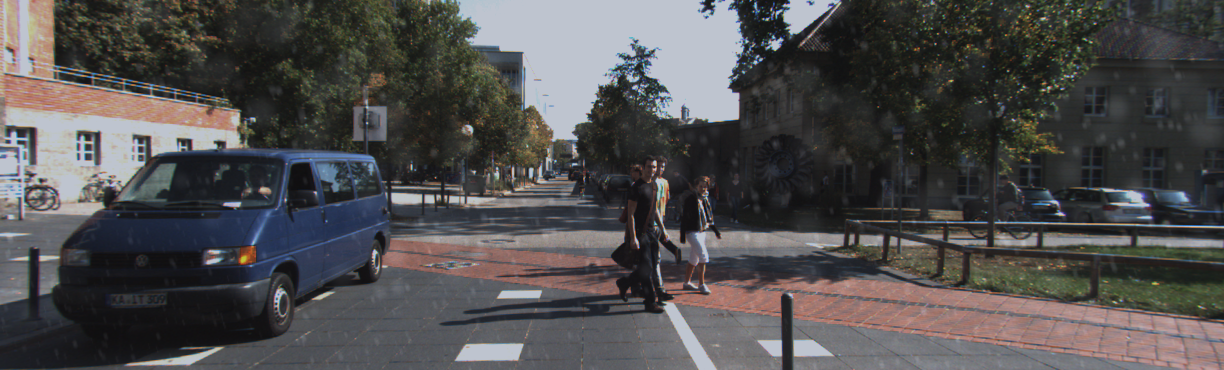}{\GSHI{0.307}}
\ImTile{SnowEight}{\XColSEight}{\YRowMid}{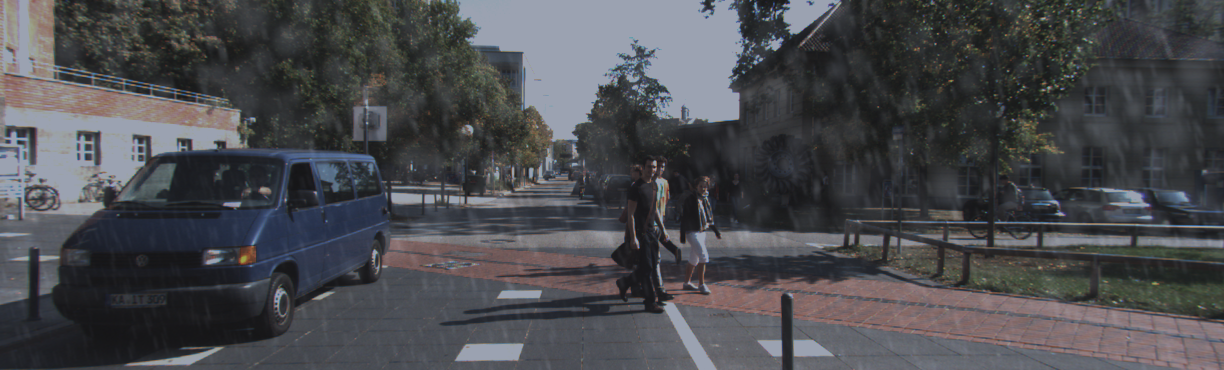}{\GSHI{0.013}}
\ImTile{SnowOne}{\XColSOne}{\YRowMid}{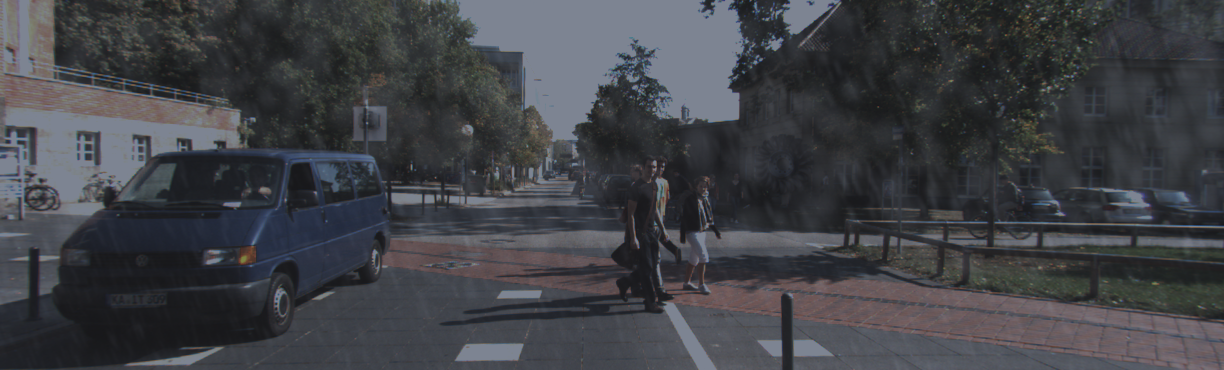}{\GSHI{0.003}}
\MapTile{SnowMap}{\XColMap}{\YRowMid}{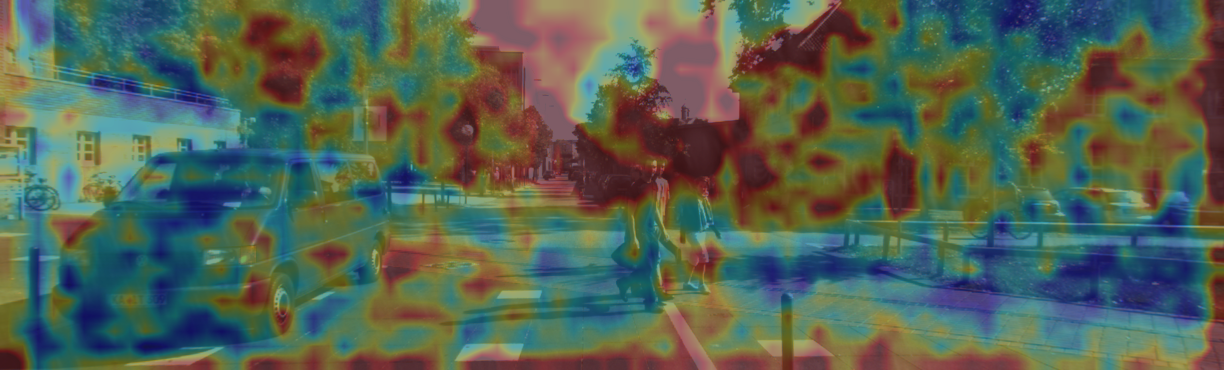}

% =========================
% Row 3: Motion Blur
% =========================
\ImTile{MBClear}{\XColClear}{\YRowBot}{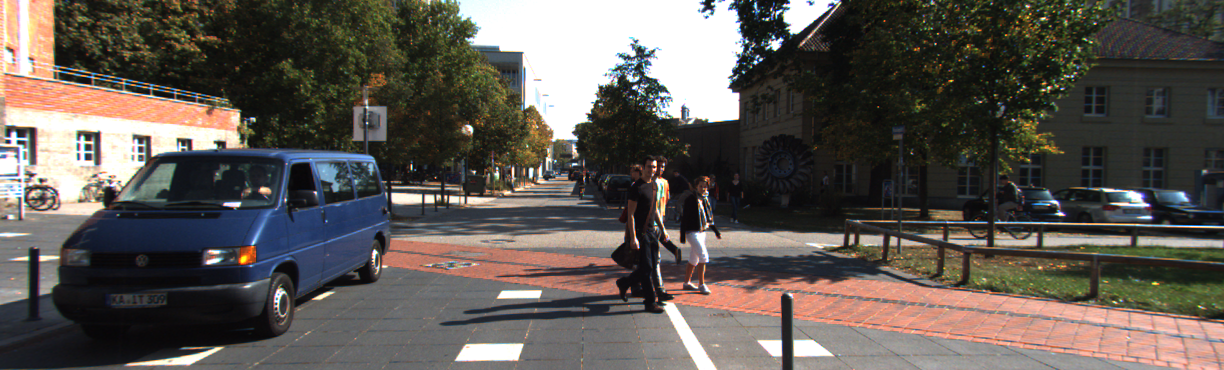}{\GSHI{0.819}}
\ImTile{MBFour}{\XColSFour}{\YRowBot}{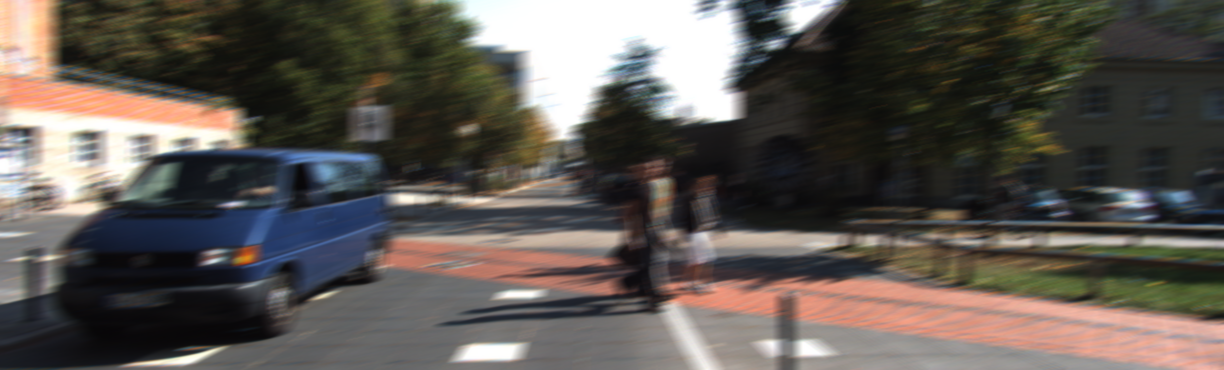}{\GSHI{0.273}}
\ImTile{MBEight}{\XColSEight}{\YRowBot}{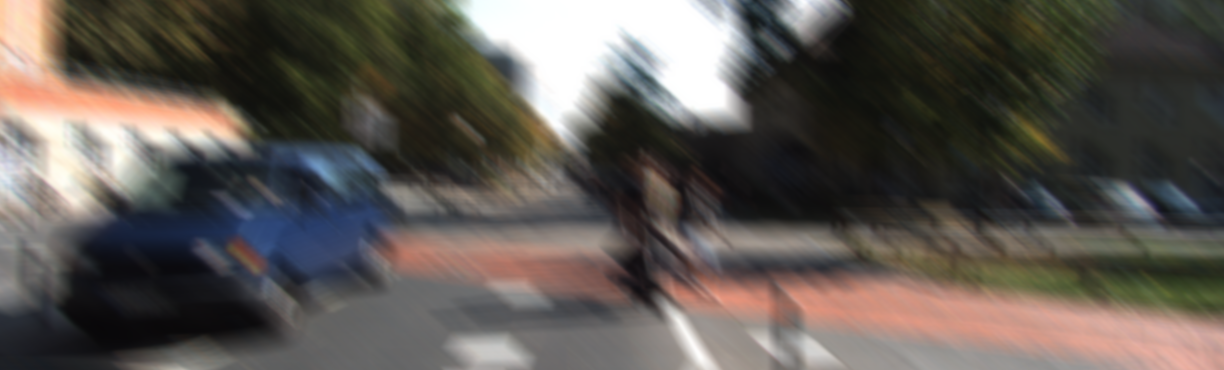}{\GSHI{0.043}}
\ImTile{MBOne}{\XColSOne}{\YRowBot}{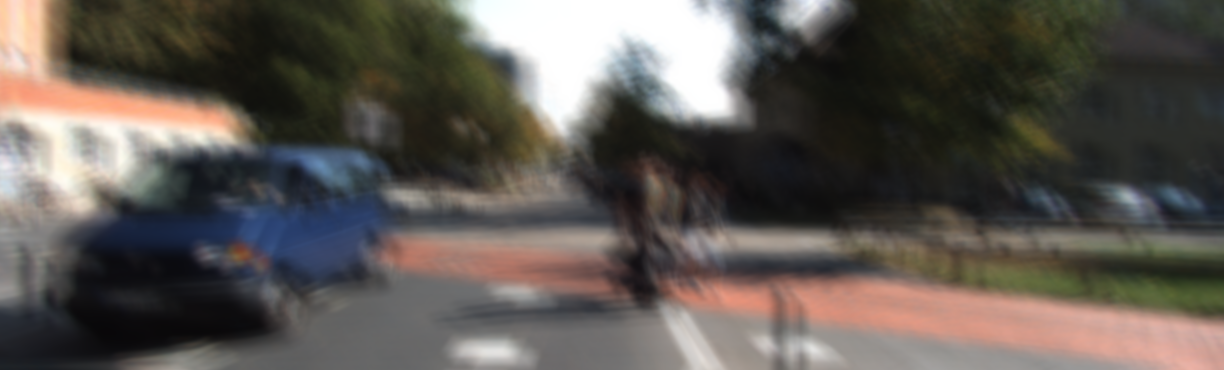}{\GSHI{0.014}}
\MapTile{MBMap}{\XColMap}{\YRowBot}{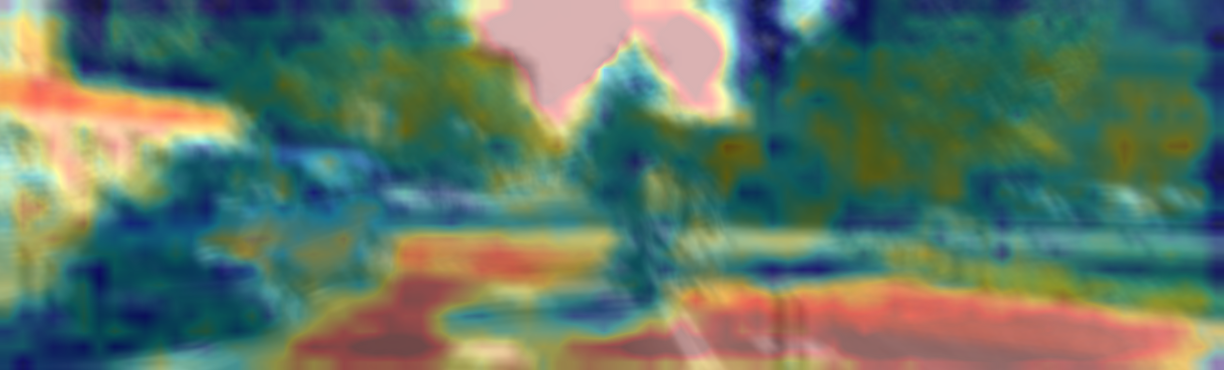}

% ---- Outer box ----
\draw[rounded corners=1pt, line width=0.5pt, gray!60]
  (\XLeftBox,\YBotBox) rectangle (\XRightBox,\YTopBox);

\end{tikzpicture}%
}

\caption{
Teaser of the proposed camera reliability monitor. As degradation severity increases, the predicted Global Sensor Health Index (GSHI) decreases while spatial uncertainty highlights unreliable image regions, providing an early-warning signal before downstream perception failure.
}
\label{fig:teaser}
\vspace{-0.5em}
\end{figure*}

\section{Introduction}

Camera-based perception is central to modern Advanced Driver Assistance Systems (ADAS), supporting object detection, lane understanding, free-space estimation, and scene-level reasoning~\cite{koopman2016challenges,burton2017safety}. In safety-critical driving, however, perception accuracy alone is an insufficient guarantee of safe operation: the system must also continuously assess whether its camera input is reliable enough to support downstream decisions. A detector may continue producing high-confidence outputs even when the image signal has been substantially degraded, deferring fallback actions until the perception stack has already entered an unsafe operating regime~\cite{hendrycks2019corruptions,michaelis2019benchmarking}.

Real-world cameras are subject to a wide range of degradations, including fog, rain, snow, low illumination, glare, motion blur, defocus blur, lens occlusion, compression artifacts, exposure shift, vignetting, and sensor noise~\cite{sakaridis2018foggy,halder2019physics,sakaridis2021acdc}. These effects frequently introduce gradual image ambiguity that predates any measurable collapse in downstream task performance~\cite{hendrycks2019corruptions,taori2020measuring}. This temporal asymmetry creates a \emph{reliability gap}: by the time task-level indicators such as detection failure, tracking instability, or cross-sensor disagreement expose the problem~\cite{zendel2018adverse,koopman2016challenges}, the vehicle may have already sacrificed valuable reaction time.

This paper addresses \emph{safety-critical camera reliability monitoring} for ADAS. Rather than treating visual degradation only as a robustness stress test~\cite{hendrycks2019corruptions,michaelis2019benchmarking} or as a generic image-quality distortion~\cite{mittal2012brisque,mittal2013niqe}, we model it directly as an indicator of perception risk. Our central hypothesis is that camera degradations induce characteristic global and spatial uncertainty patterns~\cite{kendall2017uncertainties,mcallister2017concrete} that are detectable before downstream perception failure becomes observable. Fig.~\ref{fig:teaser} illustrates this concept: as degradation severity increases, the predicted GSHI falls while spatial uncertainty maps highlight the regions of the image that have become unreliable.

We propose a degradation-aware reliability framework that jointly predicts degradation presence, continuous severity, a spatial uncertainty map, and a scalar GSHI from a single RGB image. GSHI provides a task-independent summary of camera input reliability, with values near one indicating healthy conditions and values approaching zero indicating degraded or critical states~\cite{iso26262,burton2017safety}. Unlike a simple weighted average of degradation scores, the proposed multiplicative GSHI formulation ensures that a single severe failure mode such as lens occlusion or strong motion blur can dominate the global estimate, which matches the safety intuition that one critical fault outweighs several mild distortions.

Because real driving datasets rarely provide fine-grained, per-image labels for degradation type, continuous severity, global health, and spatial degradation footprint, we train the monitor exclusively on physics- and geometry-aware synthetic supervision~\cite{sakaridis2018foggy,halder2019physics}. Twelve camera degradation modes are applied to KITTI-derived images~\cite{geiger2013kitti} with fully controlled severity parameters. For scene-dependent effects such as fog and defocus blur, monocular depth estimates~\cite{ranftl2021dpt} guide synthesis to produce spatially realistic degradations; critically, no depth information is required at inference time, so the deployed monitor takes only a single RGB image as input.

Experiments show that GSHI decreases monotonically with degradation severity and achieves a mean absolute error of $0.064$ for health estimation. Early-warning analysis shows positive lead time for seven failure-inducing degradation modes, with an average lead time of $0.47 \pm 0.25$ severity units before detector failure at $\tau_H=0.8$. A threshold-sensitivity study further shows that positive lead time is maintained across $\tau_H \in \{0.7, 0.8, 0.9\}$, indicating that the early-warning behavior is not tied to a single manually selected operating point. Zero-shot evaluation on the DAWN dataset~\cite{kenk2020dawn} demonstrates transfer to real fog, rain, and snow, and comparison against IQA~\cite{mittal2012brisque,mittal2013niqe,venkatanath2015piqe} and OOD baselines confirms the advantage of degradation-aware supervision.

The main contributions are summarized as follows:
\begin{itemize}
    \item \textbf{Proactive reliability formulation.} Camera reliability monitoring is formulated as a safety-critical perception problem for ADAS, estimating degradation-induced uncertainty \emph{before} any downstream task failure is observable.
    \item \textbf{Global Sensor Health Index (GSHI).} A risk-aware multiplicative reliability score that maps to interpretable Healthy, Degraded, and Critical operating regimes, allowing severe single-mode failures to dominate the health estimate when appropriate.
    \item \textbf{Degradation-aware synthetic supervision pipeline.} A physics- and geometry-aware synthesis framework covering twelve camera failure modes with continuous severity labels and depth-conditioned rendering for scene-dependent effects.
    \item \textbf{Lightweight multi-task monitor.} A single-image network that jointly predicts degradation type, severity, GSHI, and spatial uncertainty maps at inference, with no downstream task feedback required.
    \item \textbf{Early-warning evaluation protocol.} We introduce a detector-coupled evaluation protocol that measures whether camera-health warnings occur before downstream YOLOv8 failure, including per-degradation lead time and threshold-sensitivity analysis.
  \end{itemize}

\section{Related Work}

\textbf{Robustness under adverse visual conditions.}
The robustness of deep perception models under synthetic and real-world corruptions---including blur, noise, weather, illumination shifts, and compression artifacts---has been extensively studied. Benchmarks such as ImageNet-C~\cite{hendrycks2019corruptions} and driving-focused robustness studies~\cite{michaelis2019benchmarking,taori2020measuring} demonstrate that modern detectors and segmenters can suffer severe performance degradation under distribution shift. These studies are valuable because they expose performance cliffs that would otherwise remain invisible at deployment. However, they frame corruptions as stress tests: the goal is to measure how much downstream task performance degrades \emph{after} the perception model has already been exposed to a corrupted input. Our work takes the complementary view, treating degradation itself as a safety-relevant diagnostic signal from which perception risk can be estimated \emph{before} downstream task failure becomes observable.

\textbf{Image quality assessment.}
No-reference image quality assessment (IQA) methods such as BRISQUE~\cite{mittal2012brisque}, NIQE~\cite{mittal2013niqe}, and PIQE~\cite{venkatanath2015piqe} estimate perceptual quality from statistical image features, without requiring a clean reference. Learning-based IQA methods extend this by using deep features to regress quality scores. While these methods are effective for measuring perceptual distortion, perceptual quality is not equivalent to safety-critical perception reliability. A globally degraded image may still support accurate object detection, whereas a spatially localized artifact can compromise a safety-critical region---such as a pedestrian or a stop sign---while leaving overall image statistics largely unaffected. Furthermore, IQA methods produce a single scalar quality estimate and do not model degradation type, continuous severity, spatial uncertainty footprint, or downstream perception risk. By contrast, GSHI is designed specifically as a perception reliability score for ADAS, grounded in a degradation-aware taxonomy rather than generic visual quality statistics.

\textbf{Physics-based degradation modeling.}
Physics-based degradation models and adverse-condition datasets have been widely used for fog, rain, snow, blur, glare, noise, and compression simulation~\cite{sakaridis2018foggy,halder2019physics,sakaridis2021acdc}. Such models are essential for controlled experiments because real adverse-condition data rarely provides ground-truth severity annotations. Prior work primarily employs these models for data augmentation, image restoration, domain adaptation, or robustness benchmarking. We instead use severity-controlled, physics- and geometry-aware synthesis as a \emph{supervisory} signal for a camera reliability monitor. Depth is incorporated during synthesis for scene-dependent effects such as fog and defocus blur, using monocular depth estimates inspired by dense prediction models~\cite{ranftl2021dpt}, but the deployed monitor operates on a single RGB image without any depth input at inference.

\textbf{Uncertainty estimation.}
Uncertainty estimation in deep neural networks has been approached through Bayesian neural networks, Monte Carlo dropout, deep ensembles, and uncertainty-aware multi-task learning~\cite{kendall2017uncertainties,gal2016dropout,kendall2018multitask,mcallister2017concrete}, with applications spanning segmentation, detection, depth estimation, and out-of-distribution detection. The dominant paradigm interprets predicted uncertainty as model confidence with respect to a specific downstream task. Our work focuses instead on \emph{sensor-induced} uncertainty caused by degradation in the camera signal itself. This distinction is operationally important: a downstream detector may remain overconfident even when the image it receives has become unreliable, precisely because the model was not trained to represent the degraded input distribution.

\textbf{Sensor health monitoring.}
Sensor health monitoring in autonomous driving traditionally relies on hardware diagnostics, cross-sensor consistency checks, redundancy mechanisms, or downstream task-level failure indicators, and must ultimately be integrated with vehicle-level safety processes~\cite{zendel2018adverse,koopman2016challenges,burton2017safety,iso26262}. These approaches are valuable but inherently reactive: a fault is detected only after disagreement, instability, or accuracy loss becomes measurable. Camera-only reliability monitoring is a harder problem because degradations can be gradual, spatially localized, mixed across multiple modes, and extremely difficult to label at scale in real data. Our framework addresses this gap by learning degradation-aware uncertainty patterns directly from physics-supervised synthetic data and aggregating them into an interpretable GSHI that can provide proactive early warnings and support fallback decisions before any task-level failure is detected.

\section{Safety-Critical Camera Reliability Framework}
\label{sec:framework}

Given an RGB image $I\in\mathbb{R}^{H\times W\times3}$ captured by a forward-facing ADAS camera, the goal is to estimate input reliability \emph{before} downstream perception failure is observed. The monitor produces four outputs: degradation probabilities $\hat{\mathbf{p}}\in[0,1]^K$, severity estimates $\hat{\mathbf{s}}\in[0,1]^K$, a Global Sensor Health Index $\hat{H}\in[0,1]$, and an optional spatial uncertainty map $\hat{U}\in[0,1]^{H\times W}$:
\begin{equation}
 f_\theta(I)=(\hat{\mathbf{p}},\hat{\mathbf{s}},\hat{H},\hat{U}), \quad K=12
\end{equation}
The score $\hat{H}$ is intentionally task-independent: it quantifies camera input reliability rather than the confidence of any particular detector, 
segmenter, or tracker. This distinction is critical, because a downstream model may remain highly confident on a degraded input, especially when
 object appearance is partially preserved or when the model is poorly calibrated outside its training distribution. This distinction is consistent
  with prior observations that perception models may remain poorly calibrated under distribution shift or adverse visual conditions, 
  even when downstream predictions appear confident.

For operational use, $\hat{H}$ is mapped to three regimes: \textbf{Healthy} ($0.9\leq\hat{H}\leq1.0$), \textbf{Degraded} ($0.6\leq\hat{H}<0.9$), and \textbf{Critical} ($\hat{H}<0.6$). These thresholds define a deployment interface that can be recalibrated for a specific vehicle platform, perception stack, camera placement, operational design domain, and acceptable safety margin. In this work, the regimes are used for early-warning analysis and are not presented as universal safety limits.

\subsection{Global Sensor Health Index}

A simple weighted average of degradation severities is insufficient for safety-critical monitoring because a single severe failure mode---such as lens occlusion or strong motion blur---can render perception unsafe even when all other degradation channels are negligible. To capture this behavior, we define a multiplicative risk-aware health score:
\begin{equation}
\hat{H}(I)=\prod_{i=1}^{K}\left(1-\hat{s}_i(I)\right)^{w_i\alpha_{g(i)}}
\label{eq:gshi}
\end{equation}
where $\hat{s}_i$ is the predicted severity of degradation $i$, $w_i$ is a degradation-specific risk weight, $g(i)$ denotes the degradation group, and $\alpha_{g(i)}$ is a group-level scaling exponent. The score satisfies $\hat{H}\in[0,1]$, with values approaching one indicating healthy input.

Unless otherwise stated, the GSHI exponents use the base degradation weights and group-scale multipliers in Table~\ref{tab:gshi_weights}; the effective exponent for mode $i$ is $w_i\alpha_{g(i)}$. The effective exponent in Eq.~(2) is $w_i\alpha_{g(i)}$, where $w_i$ is the degradation-specific base weight and $\alpha_{g(i)}$ is the group-scale multiplier.

\begin{table}[t]
\centering
\caption{GSHI risk weights used for multiplicative health aggregation}
\label{tab:gshi_weights}
\begin{tabular}{lccc}
\hline
\textbf{Mode} & $\boldsymbol{w_i}$ & $\boldsymbol{\alpha_{g(i)}}$ & $\boldsymbol{w_i\alpha_{g(i)}}$ \\
\hline
Haze/Fog & 1.30 & 1.00 & 1.30 \\
Rain & 1.10 & 1.00 & 1.10 \\
Snow & 1.20 & 1.00 & 1.20 \\
Low Light & 1.30 & 1.00 & 1.30 \\
Motion Blur & 1.40 & 1.10 & 1.54 \\
Defocus Blur & 1.40 & 1.10 & 1.54 \\
Glare/Flare & 1.50 & 1.10 & 1.65 \\
Vignetting & 0.70 & 1.10 & 0.77 \\
Sensor Noise & 1.10 & 0.95 & 1.05 \\
Exposure Shift & 1.00 & 0.95 & 0.95 \\
JPEG Compression & 0.80 & 0.95 & 0.76 \\
Lens Occlusion & 1.60 & 1.15 & 1.84 \\
\hline
\end{tabular}
\end{table}

The largest effective weights are assigned to lens occlusion, glare/flare, motion blur, and defocus blur because these modes can rapidly compromise localized safety-critical evidence. Lower weights are assigned to vignetting and compression because mild forms of these degradations often preserve sufficient structure for downstream perception.

The multiplicative formulation has two safety-relevant properties. First, it is \emph{monotonic}: increasing any individual severity $\hat{s}_i$ cannot increase $\hat{H}$. Second, it is \emph{sensitive to single-mode failures}: if a high-risk degradation carries a large exponent $w_i\alpha_{g(i)}$, even moderate severity in that channel can sharply reduce $\hat{H}$ and trigger an early warning. This behavior reflects the operational safety intuition that a localized lens occlusion near a pedestrian, or a brief episode of strong motion blur, may be more perception-critical than several simultaneous mild distortions.

\subsection{Degradation Diagnosis and Severity Estimation}

Alongside the scalar health score, the monitor predicts a multi-label degradation presence vector,
\begin{equation}
\hat{\mathbf{p}}(I)=[\hat{p}_1(I),\ldots,\hat{p}_K(I)]
\end{equation}
and a per-mode severity vector,
\begin{equation}
\hat{\mathbf{s}}(I)=[\hat{s}_1(I),\ldots,\hat{s}_K(I)]
\end{equation}
These outputs make the health score interpretable and actionable. Two images may share a similar GSHI value yet require qualitatively different operational responses: a score of 0.5 caused by moderate fog calls for different fallback behavior than the same score caused by localized lens occlusion. Explicit severity estimates also allow the health score to be recalibrated offline if a deployment platform assigns different risk weights to different degradation groups.

\subsection{Spatial and Object-Level Reliability}

The spatial uncertainty map $\hat{U}$ localizes image regions that the monitor regards as unreliable. Global health scores cannot capture this: a localized artifact such as glare, vignetting, or partial lens occlusion may overlap a pedestrian or a lane boundary and compromise a safety-critical detection, even when most of the image remains visually intact. When bounding-box detections are available from a co-running detector, object-level reliability can be derived by pooling spatial uncertainty inside each box $b_j$:
\begin{equation}
 r_j = 1 - \frac{1}{|b_j|}\sum_{x\in b_j}\hat{U}(x)
\end{equation}
A low value of $r_j$ indicates that the detection lies within a spatially degraded region, providing a per-object reliability cue that complements the global GSHI.

\subsection{Fallback Interface}

The framework is designed to run as a lightweight parallel module alongside the existing perception stack. A \textbf{Healthy} GSHI supports nominal ADAS operation. A \textbf{Degraded} score can trigger cautionary behaviors such as increased following distance, reduced automation authority, or cross-sensor validation requests. A \textbf{Critical} score can initiate stronger fallback actions including driver notification, handover requests, or execution of a minimal-risk maneuver. 
The proposed monitor does not replace system-level safety validation; instead, it provides an additional diagnostic signal that can precede task-level failure and complement established safety-case and functional-safety processes~\cite{koopman2016challenges,burton2017safety,iso26262}.

\section{Degradation-Aware Synthetic Supervision}

Supervised training requires labels for degradation presence, severity, global health, and spatial degradation structure, which are difficult to obtain from real driving data, even in adverse-condition datasets designed for robust perception evaluation~\cite{sakaridis2021acdc,kenk2020dawn}. Such labels are prohibitively expensive to annotate in real driving data, and real adverse conditions are uncontrolled, may involve multiple simultaneous effects, and rarely come with continuous severity ground truth. We therefore construct a synthetic supervision pipeline in which all labels are derived analytically from the known parameters of the applied degradation operators. Each synthesized sample provides a label tuple $(\mathbf{y},\mathbf{s}^*,H^*)$ and, when applicable, a pixel-level spatial mask $M^*$.

\subsection{Degradation Taxonomy}

The taxonomy contains twelve ADAS-relevant modes:
\begin{multline}
\mathcal{D}=\{d_{\mathrm{fog}},d_{\mathrm{rain}},d_{\mathrm{snow}},
d_{\mathrm{low}},d_{\mathrm{defocus}},d_{\mathrm{motion}},\\
d_{\mathrm{glare}},d_{\mathrm{vignette}},d_{\mathrm{occlusion}},
d_{\mathrm{noise}},d_{\mathrm{jpeg}},d_{\mathrm{exposure}}\}
\end{multline}
Each mode has normalized severity $s_i\in[0,1]$. The taxonomy covers environmental conditions, optical artifacts, illumination effects, sensor and imaging-pipeline distortions, and motion-induced blur. These modes have different safety implications: fog reduces long-range visibility, motion blur destroys object boundaries, and lens occlusion can create localized zero-signal regions.

Table~\ref{tab:taxonomy} summarizes the twelve camera degradation modes used for reliability monitoring. Each mode is generated with continuous severity supervision and grouped according to its dominant physical source. This grouping is used both to organize the synthetic supervision pipeline and to define group-level risk scaling in the GSHI aggregation.

\begin{table}[t]
\centering
\caption{Camera degradation taxonomy used for reliability monitoring}
\label{tab:taxonomy}
\begin{tabular}{lll}
\hline
\textbf{Group} & \textbf{Modes} & \textbf{Reliability effect} \\
\hline
Weather & Fog, rain, snow & Visibility and contrast loss \\
Illumination & Low light, exposure & Photon starvation, saturation \\
Optical & Defocus, glare, vignette & Local/global optical ambiguity \\
Motion & Motion blur & Boundary and texture loss \\
Sensor/pipeline & Noise, JPEG & Signal and compression artifacts \\
Occlusion & Lens occlusion & Local zero-signal regions \\
\hline
\end{tabular}
\end{table}

The taxonomy covers both global degradations, such as low light, compression, exposure shift, and sensor noise, and localized degradations, such as glare, vignetting, and lens occlusion. This distinction is important because localized degradations may affect safety-critical regions even when the global image remains partially usable.

\subsection{Label Generation Protocol}

For each clean training image, a single degradation mode is sampled and assigned a severity value drawn uniformly from $[0,1]$. A fixed proportion of images is deliberately retained in their clean state, ensuring that the model learns the Healthy regime rather than treating every input as degraded. For mixed-degradation training, up to two physically compatible modes may be sampled with independently drawn severities. The resulting per-sample labels comprise a multi-hot presence vector $\mathbf{y}$, continuous severity targets for the active modes, and a scalar health target $H^*$ computed analytically from the GSHI formula. This supervision design allows the network to learn not only whether a degradation is present, but how strongly each active mode affects overall camera reliability, and how multiple co-occurring degradations interact in the multiplicative health formulation.

\subsection{Severity-Controlled Generation}

Given a clean image $I$, the synthesis operator applies one or two physically compatible degradation transforms,
\begin{equation}
 I'=T(I;d_i,s_i,\phi_i)
\end{equation}
where $\phi_i$ parameterizes stochastic rendering choices such as blur direction, flare source position, fog density, particle size, compression quality factor, or occlusion blob location. Sampling at most two degradations per image reflects the sparsity of real-world camera failures and prevents the creation of implausible combinations. Explicitly incompatible pairings are excluded: heavy rain and heavy snow are never co-applied, and extreme glare is not paired with severe low-light conditions. These constraints improve label consistency and encourage the network to learn disentangled degradation signatures that generalize more cleanly to real-world inputs.

\subsection{Physics- and Geometry-Aware Models}

Some degradations depend on scene geometry. We use monocular depth only during synthesis, not inference, using dense prediction models as relative geometric cues~\cite{ranftl2021dpt}. Let $\tilde{D}(x)\in[0,1]$ be normalized depth or disparity. Following common synthetic fog formulations~\cite{sakaridis2018foggy}, fog is generated with a Beer--Lambert-style model,
\begin{equation}
\begin{aligned}
t(x) &= \exp(-\beta(s)\tilde{D}(x)),\\
I'(x) &= I(x)t(x)+A(1-t(x))
\end{aligned}
\end{equation}
where $\beta(s)$ increases with severity and $A$ is atmospheric light. This makes distant regions more attenuated than nearby regions.

Defocus blur uses a depth-dependent circle-of-confusion proxy,
\begin{equation}
\begin{aligned}
C(x) &= \left|\frac{1}{\tilde{D}(x)+\epsilon}-\frac{1}{D_f+\epsilon}\right|,
\\
\sigma(x) &= \sigma_{\max}(s)\frac{C(x)}{\max_x C(x)}
\end{aligned}
\end{equation}
Spatially varying blur is rendered by discretizing the blur map into depth layers and compositing globally blurred images. Rain and snow are generated with streak- or particle-based rendering, consistent with prior physics-based adverse-weather rendering studies~\cite{halder2019physics}, optionally coupled with visibility attenuation. Motion blur uses trajectory-based point-spread functions. Lens occlusion is synthesized using soft refractive blobs with severity-controlled coverage. Glare and flare are generated from bright regions using bloom, halo, ghosting, and streak patterns. Vignetting is modeled as radial attenuation; sensor noise combines shot and read noise; JPEG compression, exposure shift, and low light use severity-controlled compression quality, intensity scaling, darkening, and amplified noise.

\subsection{Localized and Global Degradation Structure}

The synthetic pipeline distinguishes between global degradations and localized degradations. Global effects such as exposure shift, low light, compression, and sensor noise affect most of the image and primarily influence the global health score. Localized effects such as occlusion, glare, vignetting, and spatially varying blur can affect only selected regions. This distinction is important for ADAS because an artifact near a pedestrian, vehicle, or lane boundary may be more safety-critical than an artifact in the sky or background. The spatial masks used for pixel-level supervision are therefore generated only when the degradation has a physically meaningful spatial footprint.

\subsection{Target Health and Spatial Masks}

The target health score is computed from known severities using the same risk-aware aggregation:
\begin{equation}
H^*(I')=\prod_{i=1}^{K}\left(1-s_i^*\right)^{w_i\alpha_{g(i)}}
\end{equation}
This provides continuous supervision without manually labeling camera-health scores. The multiplicative target also allows a high-risk single degradation to produce a Critical label even when other modes are absent.

For spatially structured degradations, the simulator returns masks such as occlusion masks, flare-energy maps, vignetting attenuation fields, and fog transmission maps. Available masks are combined as
\begin{equation}
M^*(x)=\max_{i\in\mathcal{A}_{\mathrm{spatial}}}M_i(x)
\end{equation}
and used only when reliable. Samples without valid masks are excluded from pixel-level supervision to avoid noisy training. At deployment, none of these synthesis-time signals are required; the monitor receives only a single RGB image.

\section{Network Architecture and Training}

The reliability monitor uses an EfficientNet-B2 encoder~\cite{tan2019efficientnet} initialized from ImageNet-pretrained weights. This backbone offers a favorable balance between representational capacity and computational efficiency, which is important for online, real-time monitoring. For an input image $I$, the encoder extracts multi-scale feature maps $\{F_1,\ldots,F_L\}=E_\theta(I)$. A global descriptor $z=\mathrm{GAP}(F_L)$ is formed by global average pooling over the deepest feature map and fed to three lightweight prediction heads:
\begin{equation}
\hat{\mathbf{p}}=\sigma(h_{\mathrm{pres}}(z)),\quad
\hat{\mathbf{s}}=\sigma(h_{\mathrm{sev}}(z)),\quad
\hat{H}_{\mathrm{net}}=\sigma(h_{\mathrm{health}}(z))
\end{equation}
A structured health estimate $\hat{H}_{\mathrm{gshi}}$ is additionally computed from $\hat{\mathbf{s}}$ using Eq.~\eqref{eq:gshi}. Maintaining both outputs makes the health score explicitly traceable to per-degradation severity evidence rather than relying solely on a black-box scalar regression.

The direct head $\hat{H}_{\mathrm{net}}$ and the structured estimate $\hat{H}_{\mathrm{gshi}}$ serve complementary roles. The direct head can absorb calibration residuals from holistic image evidence that may not be fully captured by the per-mode severity predictions, while the structured estimate preserves interpretability by anchoring the global reliability score to the degradation diagnosis outputs. Joint supervision of both terms encourages consistency between root-cause identification and the overall health estimate, reducing the risk of a well-calibrated scalar predictor that is disconnected from the underlying degradation state.

\subsection{Spatial Uncertainty Head}

A shallow spatial head is attached to an intermediate feature map $F_m$, which retains more spatial detail than the final encoder representation. The head predicts a low-resolution uncertainty map and upsamples it to the input resolution:
\begin{equation}
\hat{U}_{\mathrm{low}}=h_{\mathrm{spatial}}(F_m),\qquad
\hat{U}=\mathrm{Upsample}(\hat{U}_{\mathrm{low}})
\end{equation}
The spatial output highlights degraded or unreliable image regions and supports the object-level reliability score in Eq.~(3). Because the spatial head is lightweight and reuses encoder features, the additional cost beyond image-level monitoring is small.

\subsection{Training Objective}

For each synthetic sample, the generation pipeline provides presence labels $\mathbf{y}$, per-mode severity targets $\mathbf{s}^*$, a scalar health target $H^*$, and an optional pixel-level mask $M^*$. The multi-task training objective combines five terms:
\begin{align}
\mathcal{L}_{\mathrm{pres}} &= \frac{1}{K}\sum_{i=1}^{K}\mathrm{BCE}(\hat{p}_i,y_i),\\
\mathcal{L}_{\mathrm{sev}} &= \frac{1}{\sum_i y_i+\epsilon}\sum_{i=1}^{K}y_i\,\mathrm{SmoothL1}(\hat{s}_i,s_i^*),\\
\mathcal{L}_{\mathrm{health}} &= \mathrm{SmoothL1}(\hat{H}_{\mathrm{net}},H^*),\\
\mathcal{L}_{\mathrm{gshi}} &= \mathrm{SmoothL1}(\hat{H}_{\mathrm{gshi}},H^*),\\
\mathcal{L}_{\mathrm{pix}} &= \mathrm{BCE}(\hat{U},M^*)
\end{align}
The full objective is
\begin{equation}
\begin{aligned}
\mathcal{L} = \;&\lambda_{\mathrm{pres}}\mathcal{L}_{\mathrm{pres}}+
\lambda_{\mathrm{sev}}\mathcal{L}_{\mathrm{sev}}+
\lambda_{\mathrm{health}}\mathcal{L}_{\mathrm{health}}\\
&+\lambda_{\mathrm{gshi}}\mathcal{L}_{\mathrm{gshi}}+
\lambda_{\mathrm{pix}}\mathcal{L}_{\mathrm{pix}}
\end{aligned}
\end{equation}
The pixel loss $\mathcal{L}_{\mathrm{pix}}$ is included only for samples with a reliable spatial mask; samples lacking a valid mask are excluded from pixel-level supervision to avoid noisy gradients. Severity regression is conditioned on the presence labels, so inactive degradation modes do not contribute to $\mathcal{L}_{\mathrm{sev}}$ and cannot corrupt the severity predictions for active modes. Loss weights $\{\lambda\}$ are set to $\lambda_{\mathrm{pres}}=1.0$, $\lambda_{\mathrm{sev}}=2.0$, $\lambda_{\mathrm{health}}=1.0$, $\lambda_{\mathrm{gshi}}=1.0$, $\lambda_{\mathrm{pix}}=0.5$ in all reported experiments.

\subsection{Online Inference}

At inference the monitor accepts a single RGB image and runs in parallel with the existing ADAS perception stack. No monocular depth, clean reference image, downstream task output, or redundant sensor is required. The four predicted outputs---degradation type, severity, GSHI, and spatial uncertainty---are all available before any downstream task failure is observed, enabling the fallback interface described in Section~\ref{sec:framework}.

\section{Experiments}

We evaluate the proposed framework from five perspectives: health-score calibration, early warning before downstream detector failure, comparison with baseline reliability signals, spatial uncertainty quality, and zero-shot transfer to real adverse weather.

\subsection{Experimental Setup and Metrics}

KITTI~\cite{geiger2013kitti} is used for controlled training and evaluation with on-the-fly synthetic degradations. DAWN~\cite{kenk2020dawn} is used for zero-shot real-world evaluation under fog, rain, and snow. The monitor uses an EfficientNet-B2 backbone and is trained with the multi-task objective in Section~V. Unless otherwise stated, a COCO-pretrained YOLOv8n detector~\cite{jocher2023ultralytics} is used as a frozen downstream detector and is evaluated consistently across all degradation types and severity levels; it is not used to train the reliability monitor. This separation ensures that the proposed monitor estimates input reliability rather than learning detector-specific failure labels.

For controlled evaluation, each validation image is degraded across a severity sweep $s\in\{0.0,0.1,\ldots,1.0\}$ for each degradation mode. This produces paired measurements of predicted health and downstream detector performance as severity increases. Clean images are retained as the Healthy reference state. For real-world evaluation, DAWN images are not fine-tuned or used for calibration; they are used only to test whether learned uncertainty patterns transfer to real adverse weather.

We report health MAE, issue mAP, severity MAE, Pearson correlation between GSHI and YOLOv8 $\mathrm{mAP}_{50:95}$, early-warning lead time, AUSE for spatial uncertainty, and runtime metrics. Health MAE evaluates global reliability calibration; issue mAP measures root-cause recognition across degradation modes; severity MAE measures continuous degradation strength; and AUSE evaluates whether predicted uncertainty aligns with spatial degradation structure. Health estimation error is
\begin{equation}
\mathrm{MAE}_H=\frac{1}{N}\sum_{n=1}^{N}|\hat{H}_n-H_n^*|
\end{equation}
Issue mAP is the mean average precision over the twelve degradation labels, and severity MAE is computed only over active degradation modes.

Early warning is defined as the severity gap between the first reliability warning and downstream detector failure:
\begin{equation}
\Delta s_{\mathrm{lead}}=s_{\mathrm{fail}}-s_{\mathrm{warn}},\quad
\hat{H}(s_{\mathrm{warn}})<\tau_H,
\end{equation}
\begin{equation}
\mathrm{mAP}(s_{\mathrm{fail}})<(1-\delta)\mathrm{mAP}(0)
\end{equation}
with $\tau_H=0.8$ and $\delta=0.20$ in the early-warning analysis. Positive lead time indicates that the monitor warns before detector failure.

\subsection{Health Calibration and Early Warning}

\begin{figure*}[!t]
    \centering
    \includegraphics[width=0.88\textwidth]{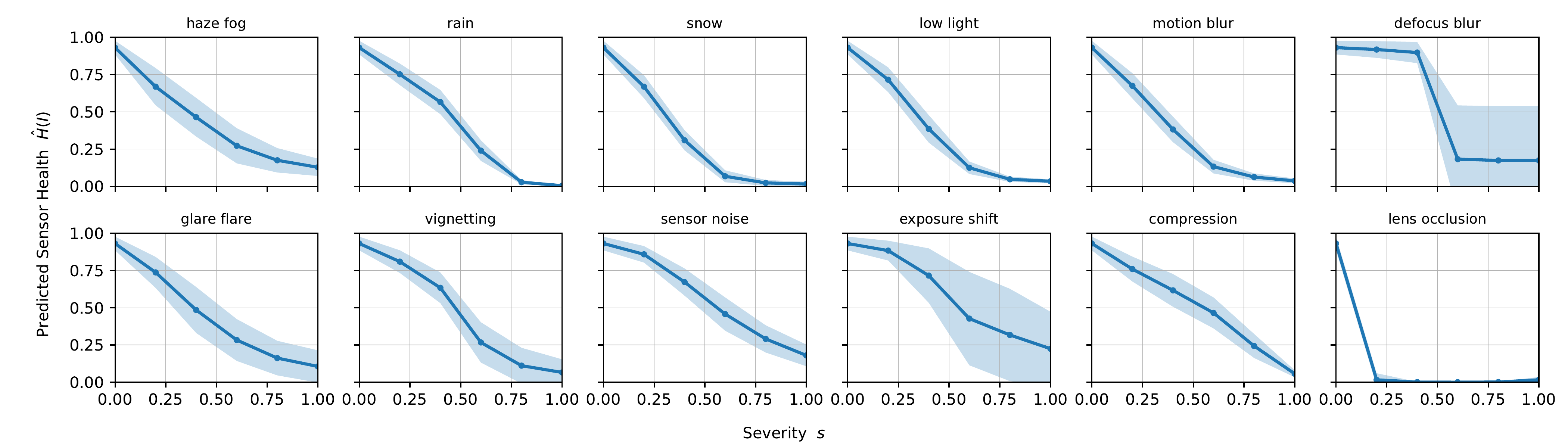}
    \caption{GSHI calibration across degradation severities. Predicted health decreases monotonically with normalized severity across twelve degradation modes, with degradation-specific response profiles that support early warning before downstream perception failure.}
    \label{fig:gshi_severity}
    \vspace{-0.8em}
\end{figure*}

Fig.~\ref{fig:gshi_severity} shows that predicted GSHI decreases monotonically with severity across all twelve degradation modes. Environmental degradations such as fog, rain, and snow show smooth health decay, while high-risk localized failures such as lens occlusion cause sharper collapse. Scene-dependent modes such as defocus, exposure shift, and glare show larger variance because their impact depends on layout, depth, and illumination. The full model achieves $\mathrm{MAE}_H=0.064$, supporting GSHI as a continuous reliability score rather than a binary failure detector. The degradation-specific response profiles are also useful for system design: gradual modes such as fog support wider warning margins, while abrupt modes such as lens occlusion or motion blur require more conservative thresholds.
To quantify whether the health estimate provides actionable warning before downstream task degradation, Table~\ref{tab:early_warning} reports per-degradation 
lead time and correlation with YOLOv8 $\mathrm{mAP}_{50:95}$. Lead time is computed as the severity gap between the first GSHI warning and the first detector
 failure point. Positive values indicate that the reliability monitor detects the degraded camera state before the detector reaches the predefined failure threshold.
GSHI provides positive lead time for seven degradation modes where YOLOv8 reaches the predefined failure threshold. 
Failure is defined as a 20\% relative drop in YOLOv8 $\mathrm{mAP}_{50:95}$; warning occurs when GSHI falls below $\tau_H=0.8$. 
N/F indicates that the detector did not reach the failure threshold within the tested severity range.

\begin{table}[t]
\centering
\caption{Early-warning and correlation analysis. }
\label{tab:early_warning}
\begin{tabular}{lcc}
\hline
\textbf{Degradation} & \textbf{Lead Time} $\uparrow$ & \textbf{Corr. w/ mAP} $\uparrow$ \\
\hline
Haze/Fog & 0.50 & 0.95 \\
Rain & 0.30 & 0.99 \\
Snow & 0.70 & 0.93 \\
Low Light & 0.30 & 0.99 \\
Motion Blur & 0.10 & 0.98 \\
Defocus Blur & N/F & $-0.91$ \\
Glare/Flare & N/F & 0.89 \\
Vignetting & N/F & 0.75 \\
Sensor Noise & N/F & 0.73 \\
Exposure Shift & N/F & 0.90 \\
Compression & 0.90 & 0.70 \\
Lens Occlusion & 0.50 & 0.25 \\
\hline
Mean over failure modes & $0.47 \pm 0.25$ & -- \\
\hline
\end{tabular}
\end{table}

\begin{table}[t]
\centering
\caption{Sensitivity of early-warning behavior to the GSHI warning threshold $\tau_H$}
\label{tab:threshold_sensitivity}
\begin{tabular}{lccc}
\hline
$\boldsymbol{\tau_H}$ & \textbf{Lead Time} $\uparrow$ & \textbf{Warning Rate} $\uparrow$ & \textbf{Trigger Rate} \\
\hline
0.7 & $0.44 \pm 0.23$ & 100.0\% & 100.0\% \\
0.8 & $0.47 \pm 0.25$ & 100.0\% & 100.0\% \\
0.9 & $0.47 \pm 0.25$ & 100.0\% & 100.0\% \\
\hline
\end{tabular}
\end{table}

The largest lead times occur for compression, snow, haze/fog, and lens occlusion, while motion blur produces the shortest lead time due to its abrupt effect on image structure. For degradation modes marked N/F, the detector did not reach the 20\% failure threshold within the tested severity range; these cases are still informative because they test whether the monitor responds to degraded sensing conditions before measurable task failure occurs.

The correlation values show that the relationship between camera health and detector performance is degradation-dependent. Weather and illumination degradations generally show strong positive correlation, while defocus blur and lens occlusion are more scene-dependent. In defocus blur, detector mAP does not decrease monotonically across the sampled severity range, producing a negative correlation despite the health monitor assigning lower reliability to stronger blur. For lens occlusion, the health score can drop rapidly because localized occlusion is treated as high risk, while global detector mAP may not decrease proportionally unless the occlusion overlaps labeled objects.

To evaluate whether early-warning behavior depends on a single threshold choice, we vary the GSHI warning threshold $\tau_H$ in Table~\ref{tab:threshold_sensitivity}. Failure is defined as a 20\% relative drop in YOLOv8 $\mathrm{mAP}_{50:95}$. Warning rate is computed over degradation modes where the detector reaches the failure threshold; trigger rate is computed over all evaluated degradation modes. Across $\tau_H \in \{0.7, 0.8, 0.9\}$, the monitor maintains positive lead time for all failure-inducing degradation modes, with mean lead time ranging from $0.44$ to $0.47$ severity units. This indicates that the early-warning result is not tied to a single manually selected operating point. In practice, $\tau_H$ should be calibrated according to the target vehicle platform, operational design domain, and acceptable false-warning rate.
\subsection{Baseline Comparison and Ablation}

We compare GSHI with no-reference image quality assessment methods (BRISQUE, NIQE, PIQE), a YOLO confidence baseline, and a clean-feature OOD baseline. IQA scores are normalized to $[0,1]$ so that larger values indicate healthier images, while YOLO confidence is computed by averaging detection confidences. The clean-feature OOD baseline extracts ImageNet-pretrained EfficientNet-B2 features, fits a clean KITTI feature distribution, and uses feature-space distance from clean images as an anomaly score. The feature extractor is frozen and receives no degradation-specific supervision or fine-tuning, making this baseline an unsupervised clean-distribution anomaly detector. These baselines test whether generic perceptual quality, detector confidence, or clean-distribution anomaly detection can substitute for a proactive camera reliability signal.
\begin{table}[t]
\centering
\caption{Comparison with baseline reliability signals}
\label{tab:baseline_comparison}
\begin{tabular}{lcc}
\hline
\textbf{Method} & \textbf{Corr. w/ mAP} $\uparrow$ & \textbf{Lead Time} $\uparrow$ \\
\hline
BRISQUE & 0.59 & No \\
NIQE & 0.73 & No \\
PIQE & 0.72 & No \\
YOLO confidence & 0.90 & No \\
Clean-feature OOD (EffNet-B2) & 0.60 & Yes, 0.19 \\
\textbf{Proposed GSHI (failure subset)} & \textbf{0.95} & \textbf{Yes, 0.47} \\
\hline
\end{tabular}
\end{table}

Table~\ref{tab:baseline_comparison} compares the proposed GSHI with generic image-quality, detector-confidence, and clean-feature OOD baselines. 
``No'' indicates that the method does not warn before the detector reaches the failure threshold. For GSHI, correlation and lead time are reported 
over the failure-inducing subset used in the baseline comparison. BRISQUE, NIQE, and PIQE show moderate correlation with downstream detector 
performance, but they do not provide reliable early warning under the selected threshold because they are not designed to estimate ADAS perception risk. 
YOLO confidence correlates more strongly with mAP, but it remains a task-dependent and reactive signal: confidence tends to degrade only after 
object evidence has already weakened. The clean-feature OOD baseline provides a more relevant anomaly-style comparison and achieves positive 
lead time of $0.19$ severity units, but its correlation remains substantially lower than GSHI. In contrast, the proposed GSHI achieves 
the strongest correlation on the failure-inducing subset and provides a larger average lead time of $0.47$ severity units before downstream perception failure.

\begin{table}[t]
\centering
\caption{Ablation study on KITTI-derived degraded images}
\label{tab:ablation}
\begin{tabular}{lccc}
\hline
\textbf{Model} & \textbf{GSHI MAE} $\downarrow$ & \textbf{Issue mAP} $\uparrow$ & \textbf{AUSE} $\downarrow$ \\
\hline
ResNet-50 baseline & 0.124 & 0.782 & 0.089 \\
EffNet-B2 global only & 0.081 & 0.854 & -- \\
\textbf{EffNet-B2 full model} & \textbf{0.064} & \textbf{0.891} & \textbf{0.042} \\
\hline
\end{tabular}
\end{table}

Table~\ref{tab:ablation} evaluates the contribution of the proposed architecture components. Lower GSHI MAE and AUSE indicate better health estimation and spatial uncertainty localization; higher issue mAP indicates better degradation recognition. Replacing the ResNet-50 baseline with EfficientNet-B2 improves health estimation and degradation recognition, reducing GSHI MAE from 0.124 to 0.081 and increasing issue mAP from 0.782 to 0.854. Adding the spatial uncertainty head further improves health estimation to 0.064 and provides localized uncertainty maps with AUSE 0.042. These results indicate that the full model benefits from both stronger global features and localized degradation evidence, supporting the design of a joint global-spatial reliability monitor.
\subsection{Spatial Uncertainty and Real-World Generalization}

\begin{figure}[t]
    \centering
    \includegraphics[width=0.92\linewidth]{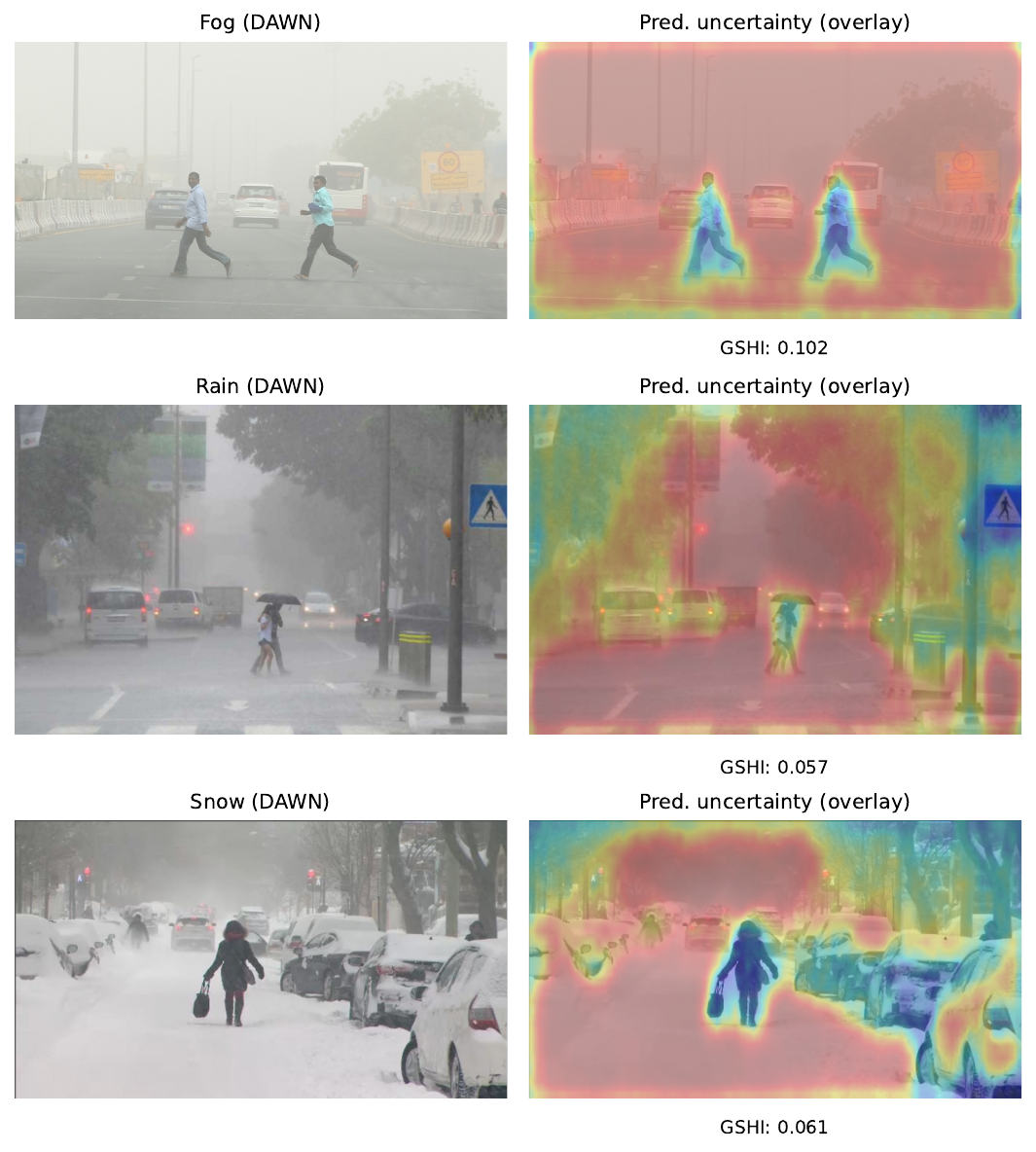}
    \caption{Zero-shot generalization on DAWN. Predicted uncertainty overlays and GSHI values under real fog, rain, and snow demonstrate transfer from synthetic training to real adverse weather.}
    \label{fig:dawn_generalization}
    \vspace{-0.8em}
\end{figure}

The spatial uncertainty maps localize degraded regions rather than acting as generic saliency maps. For degradations with reliable masks, the full model achieves AUSE $0.042$ compared with $0.089$ for the ResNet-50 baseline. This indicates that the learned uncertainty aligns with physically degraded regions and can provide interpretability for the global health score.

When evaluated on DAWN without fine-tuning, the model identifies real fog, rain, and snow with balanced accuracy of $84.2\%$. Fig.~\ref{fig:dawn_generalization} shows that low GSHI values and structured uncertainty overlays are produced under real adverse weather, supporting synthetic-to-real transfer of degradation-induced uncertainty patterns. Fog produces broad uncertainty in low-visibility regions, rain increases uncertainty around streaks and low-contrast objects, and snow emphasizes particle-heavy regions and obscured object boundaries.

This evaluation is intentionally zero-shot: no DAWN image is used for training or threshold tuning. The result therefore tests whether the model has learned general uncertainty patterns associated with physical degradation rather than memorizing synthetic artifacts. The low GSHI values in Fig.~\ref{fig:dawn_generalization} should be interpreted as conservative reliability estimates under adverse weather. Such behavior is useful for safety monitoring because the goal is to flag degraded sensing conditions early, even when a downstream detector may still produce outputs.

\subsection{Runtime and Deployment}

The monitor is designed as a lightweight parallel module that receives the same RGB input as the ADAS perception stack and does not require detector outputs, temporal history, redundant sensors, or depth at inference. Its outputs can support warning, confidence modulation, cross-sensor validation, maintenance alerts, or fallback triggering. Object-level reliability is computed by pooling $\hat{U}$ inside detector boxes and therefore adds negligible cost.

\begin{table}[t]
\centering
\caption{Runtime and deployment analysis measured at input resolution $224 \times 224$ on an NVIDIA GeForce RTX 5090}
\label{tab:runtime}
\begin{tabular}{lcccc}
\hline
\textbf{Model} & \textbf{Params} & \textbf{FLOPs} & \textbf{Latency} & \textbf{FPS} \\
 & \textbf{(M)} & \textbf{(G)} & \textbf{(ms)} & \\
\hline
ResNet-50 baseline & 25.99 & 4.23 & 1.43 & 701.4 \\
EffNet-B2 global only & 7.74 & 0.68 & 2.22 & 449.4 \\
EffNet-B2 full model & 7.91 & 0.72 & 2.27 & 440.5 \\
\hline
\end{tabular}
\end{table}

As shown in Table~\ref{tab:runtime}, the full model adds only 0.17M parameters and 0.04G FLOPs over the global-only EfficientNet-B2 variant, while maintaining real-time throughput above 400 FPS on the measured GPU. Latency and FPS are averaged over 300 forward passes after 50 warm-up iterations with batch size 1. This indicates that the spatial uncertainty head introduces minimal overhead relative to the backbone and can run in parallel with the downstream perception stack.

Although EfficientNet-B2 has substantially lower parameter count and FLOPs than the ResNet-50 baseline, measured latency depends on hardware-specific kernel efficiency. The full model nevertheless runs at 440.5 FPS, supporting online deployment.

In deployment, GSHI thresholds should be calibrated using platform-specific data, operational speed, sensor redundancy, and acceptable safety margins. For resource-constrained systems, the spatial head can be disabled and only image-level reliability outputs can be used; for higher-performance systems, both global and spatial outputs can support more detailed confidence modulation.

\section{Failure Cases and Limitations}

Several limitations of the current framework merit attention. First, physics- and geometry-aware synthesis improves training realism but cannot fully reproduce the complexity of natural degradation mixtures, such as simultaneous windshield contamination, road spray, reflections, exposure instability, and motion blur encountered in real driving. The monitor may therefore produce conservatively low GSHI estimates on unfamiliar real-world combinations. While conservatism is preferable to missed faults in safety-critical settings, excessive false warnings can erode system availability and driver trust.

Second, real driving datasets rarely provide continuous degradation severity annotations or ground-truth camera health labels. The DAWN evaluation validates synthetic-to-real transfer qualitatively and through weather recognition accuracy, but it does not constitute a quantitative calibration of GSHI against real-world ground truth. A rigorous real-world validation would require a controlled data collection protocol in which visibility, lens contamination level, illumination, and downstream perception performance are measured jointly and continuously.

Third, the Healthy, Degraded, and Critical regime thresholds (0.9 and 0.6) are not universal safety limits. They must be calibrated for each vehicle platform, camera mounting position, perception stack, operational design domain, and required safety margin. A high-speed highway system will require more conservative thresholds than a low-speed parking-assistance system.

Fourth, the twelve-mode taxonomy does not cover all possible camera faults. Rolling-shutter distortion, wiper occlusion, lens misalignment, dead pixels, thermal artifacts, color-channel failure, and long-term sensor aging are absent. Unseen degradation modes may produce elevated uncertainty estimates, but root-cause identification will be unreliable.

Fifth, the monitor is single-frame. Temporal integration could suppress false positives from transient objects, help distinguish persistent lens contamination from momentary occlusion, and improve sensitivity to slowly evolving degradations such as gradual fogging or progressive wiper smear. Sixth, the object-level reliability score is a simple spatial pooling operation. Future extensions could weight individual spatial locations by object class, distance, motion state, and driving context to better assess whether localized uncertainty is safety-critical for the current scene.

Finally, controlled evaluation in this work is limited to KITTI-derived degradations. Future work will assess cross-dataset calibration on additional benchmarks such as ACDC~\cite{sakaridis2021acdc}, BDD100K~\cite{yu2020bdd100k}, nuScenes-C~\cite{caesar2020nuscenes}, and Waymo-derived~\cite{sun2020waymo} adverse-condition splits to test generalization more comprehensively. The proposed GSHI should be regarded as a complementary diagnostic input to a broader safety architecture, not as a stand-alone functional-safety mechanism.

\section{Conclusion}

This paper presented a proactive, safety-critical camera reliability monitor for ADAS perception. The framework models camera degradation as a direct perception risk and jointly predicts degradation type, severity, spatial uncertainty, and a Global Sensor Health Index from a single RGB image, with no downstream task feedback required at inference. The risk-aware multiplicative GSHI formulation ensures that severe single-mode failures dominate the reliability estimate and maps naturally to interpretable Healthy, Degraded, and Critical operating regimes. A physics- and geometry-aware synthetic supervision pipeline covering twelve camera failure modes provides fully labeled training data without manual annotation.

Experiments demonstrate that GSHI decreases monotonically with degradation severity, achieves a health-estimation MAE of 0.064, and provides positive early-warning lead time of $0.47 \pm 0.25$ severity units for seven failure-inducing degradation modes. Threshold-sensitivity analysis confirms that this early-warning behavior is stable across $\tau_H \in \{0.7, 0.8, 0.9\}$. 
Ablation experiments validate the contribution of the EfficientNet-B2 backbone and spatial uncertainty head. Comparison with IQA methods, detector-confidence signals, and a clean-feature OOD baseline shows that the proposed GSHI achieves superior correlation with downstream detection performance and substantially longer warning lead time than all alternatives. Zero-shot evaluation on DAWN confirms that learned uncertainty patterns transfer to real fog, rain, and snow without any fine-tuning.

Future work will extend the framework to temporal sequence monitoring, broader degradation taxonomies covering unseen fault types, open-set camera anomaly detection, multi-sensor reliability fusion, cross-dataset calibration on benchmarks such as ACDC and nuScenes-C, and platform-specific threshold calibration for production ADAS deployments.

% Use BibTeX for the final paper if possible:
\bibliographystyle{IEEEtran}
\bibliography{references}

\end{document}